# Maintenance automation: methods for robotics manipulation planning and execution

Christian Friedrich, Ralf Gulde, Armin Lechler, and Alexander Verl

*Abstract*— **Automating complex tasks using robotic systems requires skills for planning, control and execution. This paper proposes a complete robotic system for maintenance automation, which can automate disassembly and assembly operations under environmental uncertainties (e.g. deviations between prior plan information). The cognition of the robotic system is based on a planning approach (using CAD and RGBD data) and includes a method to interpret a symbolic plan and transform it to a set of executable robot instructions. The complete system is experimentally evaluated using real-world applications. This work shows the first step to transfer these theoretical results into a practical robotic solution.**

*Note to Practitioners*— **This paper is inspired by the actual deficit to automate maintenance tasks in manufacturing environments through robotic systems. The focus of this work is on providing a practical robotic solution which can automate maintenance tasks completely autonomously. For this, the different skills from manipulation planning to the control and execution are explained. One main aspect of this work is to show how to create elementary robot motion tasks from symbolic plans. The second important part is to explain the practical capabilities applied on real-world applications.**

## I. INTRODUCTION

Maintenance automation is a relatively new research field with a significant potential with regard to economic objectives. In case of machine tools, for example, up to 35% of the life cycle costs are attributable to maintenance [1]. A relevant point to automate these tasks is the autonomous planning of the disassembly and assembly sequence as well as the execution under real-world conditions using a robotic system. This includes especially computational efficient algorithms as well as to incorporate environmental uncertainties in task planning. Today, there is no practical solution known for maintenance automation. Developing new methods, i.e. allowing robots the execution of repair and service tasks automatically, can be a small step towards solving these future challenges in manufacturing automation.

This paper is structured as follows: Section II discusses the state of the art in maintenance and disassembly and assembly automation. In section IV, the preliminary work for manipulation planning for maintenance automation is briefly discussed. Section V presents a new concept for transforming symbolic plans into executable robot programs. The used robot control interface as well as the different controller approaches are experimentally validated in section VI. A detailed experimental validation of the real-world applications are shown in section VII. In the last section, relevant future research topics are discussed.

## II. STATE OF THE ART

This section presents known methods for maintenance automation and the related field of (dis-)assembly automation and discuss their deployment in robotic systems.

### A. Maintenance automation

Solutions for fault diagnosis [1] and condition monitoring [33] are well known for maintenance applications, but today mainly a human operator performs the fault corrections. The use of robotic solutions for these applications is mostly known for areas that are inaccessible to humans. Current systems are especially designed for the use in specific areas like energy transmission networks [3], [4], sewer systems [5], bridges [6], or forestry [32]. Also for complex areas like material flow systems [7] or process plants [8], [9] the use of robotic systems was investigated, which is focused mainly on inspection tasks. In addition, some service and repair tasks are developed, but the task and the path were manually programmed. The scheduling of maintenance tasks is already supported by adaptive strategies, for example see [31].

To increase the flexibility and for planning under environmental uncertainties the predefinition of the robot program is not very recommendable.

### B. Related fields: (dis-)assembly automation

Opposed to maintenance the field of (dis-)assembly automation is well studied in the state of the art. Based on the requirements the planning process in assembly automation is executed using CAD assembly models. One of today's most advanced systems is described in [10], [11]. For generating a robot program automatically, an efficient algorithm [12] is developed for computing a geometrically feasible assembly sequence. The robot program is described by the concept of skill primitives [13], which provides a well-suited method for coupling planning with execution. The system is further extended in [14] with a grasp planner, whereby the autonomy is further increased. Because no sensor data is integrated into the planning process, it is not possible to incorporate environmental uncertainties. In addition, the planning time is relatively long, because the sequencing process is focused on finding a nearly-optimal assembly sequence. There exist also efficient planning algorithms which allows a dual-optimization [15] or sequencing methods which handles uncertainty, using fuzzy

Manuscript received January 01, 2022. Research supported by the German Research Foundation (DFG). Christian Friedrich is with the HKA Karlsruhe, Moltkestraße 30, 76133 Karlsruhe. Ralf Gulde, Armin Lechler, and Alexander Verl are with the Institute of Control Engineering of Tool

Machines and Manufacturing Units (ISW), University Stuttgart, Seidenstraße 36, 70174 Stuttgart, Germany (corresponding author: Christian Friedrich; phone: +49 721 925 1723; fax: +49 721 925 1707; e-mail: christian.friedrich@h-ka.de).

theory [16]. In disassembly automation [17], mainly applied to recycling tasks, the combination of CAD and vision data is very common [18], [19], because the real product differs in most cases from the prior model. Also, there exist fast disassembly planning algorithms with a two-pointer detection strategy which are evaluated in a simulation study [34]. In [35] they introduce a flexible sequencing approach to handle uncertainties. But in these works, no general method is provided which allows task and path planning in an embedded scheme. A further drawback is also that no approach is described to fuse the prior known CAD model with the information from vision in a suitable manner.

### C. Contribution

This work presents a robotic system which can automate advanced maintenance tasks with the focus on disassembly and assembly operations (no repair tasks) under real-world conditions. The main novelty of this work is the integration of the developed algorithms from preliminary work (section IV) in a complete robotic system. The focus is that we show experiments and not only simulations for the disassembly and assembly operations. We present the performance of our system (planning and execution) using real-world examples. Beyond this, we will explain a straightforward method for the generation of elementary robot motion tasks from symbolic plans. This is unique, because this concept helps us to translate the high-level plans into reactive skill primitives in a flexible way. We hope that this paper will inspire to bring the planning algorithms in real-world robotic applications.

## III. NOTATIONS

To support the readability of this paper we summarize and describe the most relevant symbols in the following table.

TABLE I. NOTATIONS

| Scalars and symbol | Explanation |
|---|---|
| $\mathcal{SSR}$ | Symbolic spatial relation |
| $\mathcal{FG}$ | Feature geometry |
| $s d o f$ | Symbolic degree of freedom |
| $\mathcal{MP}$ | Manipulation primitive |
| $\mathcal{AP}$ | Skill primitive |
| $\mathcal{HM}$ | Hybrid move, describes the robot motion |
| $\mathcal{TF}$ | Task frame; reference coordinate system |
| $\mathcal{C}$ | Component |
| $\tau$ | Tool command |
| $\lambda$ | State transition |
| **Vectors and matrices** | **Explanation** |
| $\underline{u}$ | Velocity control command |
| $\underline{R}_A$ | Admittance controller |
| $\underline{R}_{IBVS}$ | Image based visual servoing controller |
| $\underline{R}_p$ | Position controller |
| $\underline{F}$ | Force and torque |
| $\underline{M}$ | Mass |
| $\underline{C}$ | Stiffness |
| $\underline{D}$ | Damping |
| $\underline{J}_f$ | Feature jacobian |
| $\underline{f}$ | Image feature |
| **Spaces and sets** | **Explanation** |
| $\mathbb{W}_D$ | Disassembly space |
| $\Sigma$ | Symbolic actions, here $\mathcal{MP}$ |
| $\mathcal{S}$ | Uniform sampled sphere |

The symbolic spatial relations describe in a symbolic way how the components are geometrically connected on a selected feature geometry. The feature geometry defines the geometric element like a plane, line or a point. The model and the used features are based on the work from [24]. Using this information, we can compute a disassembly space which contains the geometric possibilities to move a component, see IV-A. The results are stored in a symbolic way, using an undirected graph where the components are the nodes and the vertices are the symbolic degree of freedom. This enables us to consider the planning problem in a symbolic way. The different frames and robot motions are described in V-A. The details about the controllers are shown in VI.

## IV. PRELIMINARY WORK

In this section, the preliminary work is explained for an easier understanding of the complete system. The main functionalities are divided into a task planner, an environment model and task exploration, a path planner and an interpreter for generating the robot program, which is then executed in a hybrid control architecture. An overview of the processing pipeline is presented in Fig. 2. The details of the different algorithms are already described in the prior publications [20], [21] for task and path planning and [22], [23] for task exploration. This paper discusses further the steps of robot program generation as well as the Cartesian control interface.

### A. Task planning

The task planner, for details see [20], [21], creates a symbolic plan starting from CAD assembly models and/or basic contact information from vision data. It is divided into a pre-processor for computing the relative disassembly space for every component and a processor for finding an adequate plan for the disassembly task. Computing the relative disassembly space is based on the relational assembly model suggested by [24]. It contains, among others, the contact information described by a set of symbolic spatial relations $\mathcal{SSR} \in \{concentric, congruent, screwed, \ldots\}$ as well as the feature vector $\underline{f}_{\mathcal{SSR}}$ on the connected feature geometry $\mathcal{FG} \in \{plane, line, \ldots\}$. A potential disassembly space can this way be found by a very efficient sampling-based approach. For this, a uniform sampled sphere $\mathcal{S} = \left\{\underline{x} \in \mathbb{R}^3 \mid \|\underline{x}\|_2 = 1\right\}$ is used for the representation of the initial relative disassembly space $\mathbb{W}_D$ (all normal vectors on $\mathcal{S}$ represent possible translational disassembly directions). Further, for every defined contact the corresponding disassembly space is defined. For example, the disassembly space of a plane contact forms a hemisphere, compare Fig. 1.

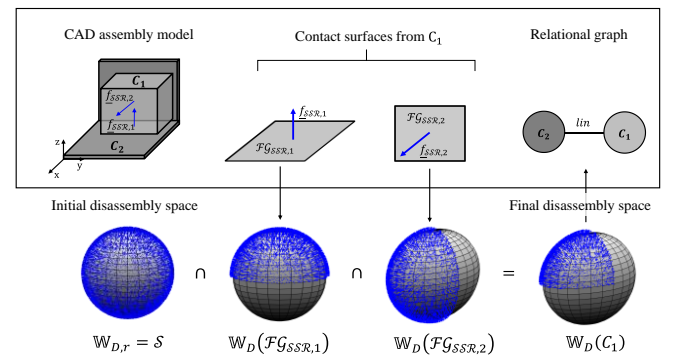

Fig. 1. Pre-processing approach for the computation of the dis-assembly space for component $C_1$.

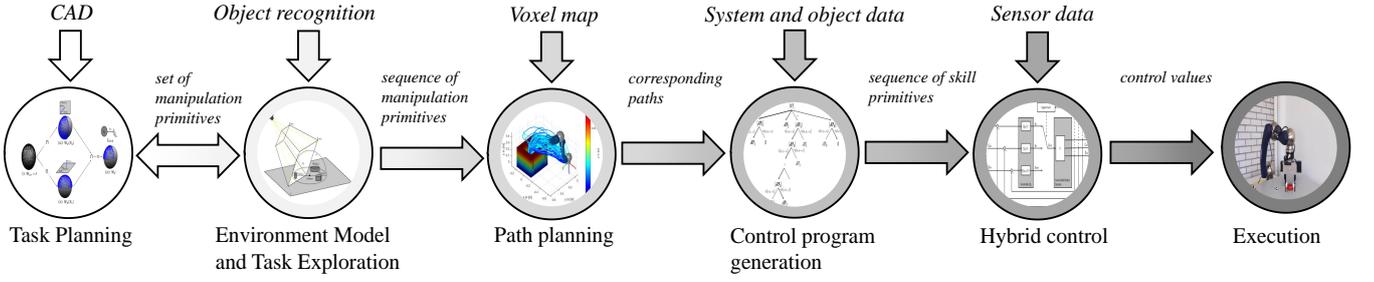

Fig. 2. System overview for the proposed manipulation planning approach.

Now the full disassembly space for a component $C_j$ is (1).

$$\mathbb{W}_{D,j} = \left( \bigcap_{i=1}^{M} \mathbb{W}_D \big( \mathcal{FG}_{\mathcal{SSR},i} \big) \right) \cap \mathbb{W}_D. \qquad (1)$$

Evaluating (1) is, due to the discrete representation, very efficient, as it can be solved using sort and compare operations. Therefore, the complexity of the proposed algorithm is $O(n \cdot log(n))$ for $n$ initial normal vectors. The complete procedure is explained in Fig. 1.

Using the computed disassembly space a relational graph $\mathcal{G}_{sdof}(C, sdof)$ can be derived, where the components $C$ represent the nodes and the edges are described by $sdof \in \{fix, lin, rot, fits, agpp, \ free\}$ which decode the relative degree of freedom in a symbolic manner [24].

Defining a set of symbolic actions $\Sigma$, further referred to as manipulation primitive (we use this to describing purely symbolic actions) $\Sigma = \{\mathcal{MP}\} = \{move, twist, put, pull\}$ allows the description of a state transition $\delta \colon \mathcal{Z} \times \Sigma \to \mathcal{Z}$, for a set of states $\mathcal{Z} := sdof_j(C_i), \forall i \in \mathcal{G}_{sdof}$, with the state transition $sdof_{k+1}(C_i) := \delta(sdof_k(C_i), \mathcal{MP}_k)$. The used tool described by $\tau_j$ can be inferred by the components semantic and the selected manipulation primitive. Updating $sdof_{k+1}$ is done by a rule-based approach or, if no rule is applicable, through a new query to the pre-processor.

### B. Environment, task exploration and path planning

To build up an environmental model and to handle uncertainty, we are using a vison system [22] and a next-best-view algorithm [23]. The path planning works on the basis of a voxel map and global search- and sampling-based planning algorithms. The sequencing of the manipulation primitives is described as a traveling salesman problem. For details see [20], [21].

### C. Practical implementation

The complete system is implemented in ROS [25]. The different functions are developed in Matlab and C/C++. Hardware interconnection with the Schunk LWA4p robot is created via the CANopen stack [26]. As sensors a 6D force-torque-sensor (Schunk FTM 75) and a Microsoft Kinect v2 are used. The system runs on an Intel Core i7-4790K, 4000MHz with 16GB DDR3 RAM using Linux Ubuntu.

## V. FROM SYMBOLIC PLANS TO SKILL PRIMITIVES

This section proposes a solution to interpret symbolic plans, translating manipulation primitives directly into skill primitives, which can be executed on the robot controller. Therefore, a general sequence of skill primitive is predefined, which can individual adapted to the actual states. This allows us to build a flexible solution which can be parameterized by the planning system.

### A. Skill primitives

To define the control program the formalism of skill primitives is used, because it provides a well-suited method to describe elementary robot motions in a hybrid control scheme. Further, we introduce the most relevant details of the approach to explain our method in the next section. The complete framework is described in [13], [27]. A skill primitive $\mathcal{AP}$ is defined according to (2).

$$\mathcal{AP} := \langle \mathcal{HM}, \tau, \lambda \rangle. \qquad (2)$$

Here the hybrid move $\mathcal{HM}$ describes the robot motion in a defined coordinate system and the used control method. In our approach, $\mathcal{TF} \in \{world, tcp, tool_i, rgbd\}$ is used as reference coordinate system. As $control$ a position $pos$, a force-torque $ftc$ and a visual-servoing controller $vsc$ is used. $\tau := \langle tool, cmd \rangle$ describe the available $tools \in \{gripper, screwdriver\}$ and the corresponding commands. The skill primitive is terminated by the stop condition $\lambda \colon \mathcal{S} \to \{true, false\}$ for a predefined set of sensor conditions.

### B. Plan decomposition

To generate an executable control program, we decompose the symbolic plan into low level skill primitives. The symbolic plan, coming from the task planner is described in a linear sequence of manipulation primitives (3) as mentioned in IV-A.

$$\langle \mathcal{MP}_0 \big( C_i, \tau_j \big), \dots, \mathcal{MP}_{N-1}(C_k, \tau_l), \mathcal{MP}_N(C_m, \tau_n) \rangle \qquad (3)$$

This describes the abilities of the robot like $\{move, twist, put, pull\}$ which are executed on a component $C$ using a tool $\tau$. Further we will explain how to parse this plan in an executable robot program described in skill primitives. The decomposition of the manipulation primitive depends on the used tool, the applied component and the next manipulation primitive in the plan. A sequence that describes the process-dependent set $\mathcal{MP}^C = \{\mathcal{MP}\} \backslash \{move, put\}$ can be specified with (4)

$$\begin{aligned}
\mathcal{MP}^C(\forall \tau, \forall C) \to [&getTool_i] \quad [getObj] \\
&move \quad \{processObj_1\} \\
&\{processObj_2\} \quad \dots \\
&\{processObj_n\} \quad move \\
&put \quad [putTool_i].
\end{aligned} \qquad (4)$$

In this case, $[x]$ refers to a call to $x$ at most once, and $\{x\}$ represents a repetition of the skill primitive at least once. For the $move$-primitve the decomposition (5) and for the $put$-primitive the decomposition (6) are defined.

$$move(\forall \tau, \forall C) \to [roughPos] \quad [finePos]. \qquad (5)$$

$$put(\forall \tau, \forall C) \to [putObj_j] \quad [move]. \qquad (6)$$

The skill primitives for $processObj_i$ contains different steps to solve (e.g. a screwing application, see Fig. 11). As an example, a rough positioning (7) describes a position-controlled movement which can be supplemented by a fine positioning (8) using visual-servoing control. The two-stage approach movement, consisting of rough and fine positioning, is used to compensate for the inaccuracy of the pose estimation of the object localization.

$$
roughPos \begin{cases} \mathcal{HM} \begin{cases} \mathcal{D} \begin{cases} \mathcal{TF} = world \\ control = (pos \quad ... \quad pos) \\ d = (p_x \quad p_y \quad p_z \quad \alpha_x \quad \beta_y \quad \gamma_z) \end{cases} \\ \tau = \begin{cases} tool =/ \\ cmd =/ \end{cases} \\ \lambda = \underline{p}_g \end{cases} \end{cases} \quad (7)
$$

$$
finePos \begin{cases} \mathcal{HM} \begin{cases} \mathcal{D} \begin{cases} \mathcal{TF} = rgbd \\ control = (vsc \quad ... \quad vsc) \\ d = (\hat{p}_{1,u} \quad \hat{p}_{1,v} \quad ... \quad \hat{p}_{3,u} \quad \hat{p}_{3,v}) \end{cases} \\ \tau = \begin{cases} tool =/ \\ cmd =/ \end{cases} \\ \lambda = \underline{\hat{f}}_g \end{cases} \end{cases} \quad (8)
$$

### C. Decision rules

For the resolution of the individual skill primitive calls, decision-making rules $r(c_1, c_2, ..., c_j) \to \{true, false\}$ are introduced with the condition $c_i$. Means, it is a pure evaluation of the boolean expression. This allows the general sequence (4)-(6) to be transferred into the respectively required one. The following rules (9)-(14) are sufficient for deciding whether a skill primitive is inserted or not into the executable sequence. $\underline{p}_{g,i}$ is the current goal, $\underline{p}_{vp}$ the position from visual servoing and $\underline{p}_{robot}$ the current robot position.

$$getTool \leftarrow (tool_i \neq tool_{i-1}) \vee (tool_i = \emptyset) \quad (9)$$

$$putTool \leftarrow (tool_{i+1} \neq tool_i) \vee (\mathcal{MP}_{i+1} = \emptyset) \quad (10)$$

$$roughPos \leftarrow (\underline{p}_{g,i} = \underline{p}_{robot}) \quad (11)$$

$$finePos \leftarrow (\underline{p}_{g,i} = \underline{p}_{vp}) \quad (12)$$

$$putObj_j \leftarrow (C_{i+1} \neq C_i) \quad (13)$$

$$getObj \leftarrow (\mathcal{MP} \in assemblySequence) \quad (14)$$

Fig. 3 explains the procedure of the rule-based decomposition. Only if the conditions $r_1(c_1, c_2, ..., c_i)$ for $[\mathcal{AP}_1]$ are satisfied, this is executed. If $r_3(c_1, c_2, ..., c_j)$ is satisfied $\{\mathcal{AP}_3\}$ is repeated until $\neg r_3(c_1, c_2, ..., c_j)$ holds. Then, the next skill primitive, which is given from the sequence (4)-(6), is processed further. This ruled-based approach allows the decomposition of the symbolic plan in a sequence of executable skill primitives during the operation time. Because there are only a few states which must be considered, the decomposition is very efficient and due to the symbolic description very flexible to describe robot manipulation tasks.

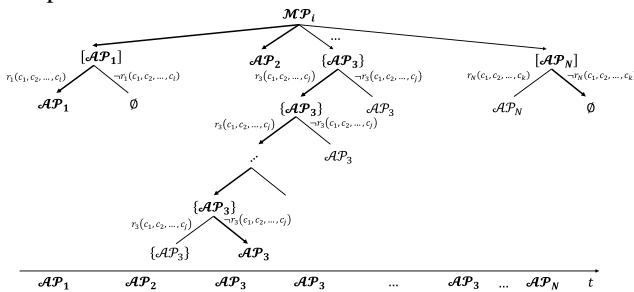

Fig. 3. Decision tree to transfer a symbolic manipulation primitive into a sequence of executable skill primitives.

### D. Runtime interpretation

The decomposition, discussed in the previous section, and the examination of the decision-making rules for determining the skill primitives to be called are carried out by an interpreter at runtime. This permits a flexible and fault tolerant execution of the manipulation primitives, since in the event of a task failure, an alternative plan can be directly determined and executed.

The real-time layer embeds the task controller and sends Cartesian velocity commands to the robot controller via the robot interface. The robot interface manages the robot-states (has gripper, has object, ...) and object-states (component position after the execution of a $put$-primitive, ...). After an execution of a skill primitive the actual states are updated for an evaluation of the decision rules (9)-(14).

Fig. 4 shows the signal flow of the approach to generate the executable robot program. After the decomposition of the different manipulation primitives, a set of skill primitives is generated. These are further parameterized by the planning system.

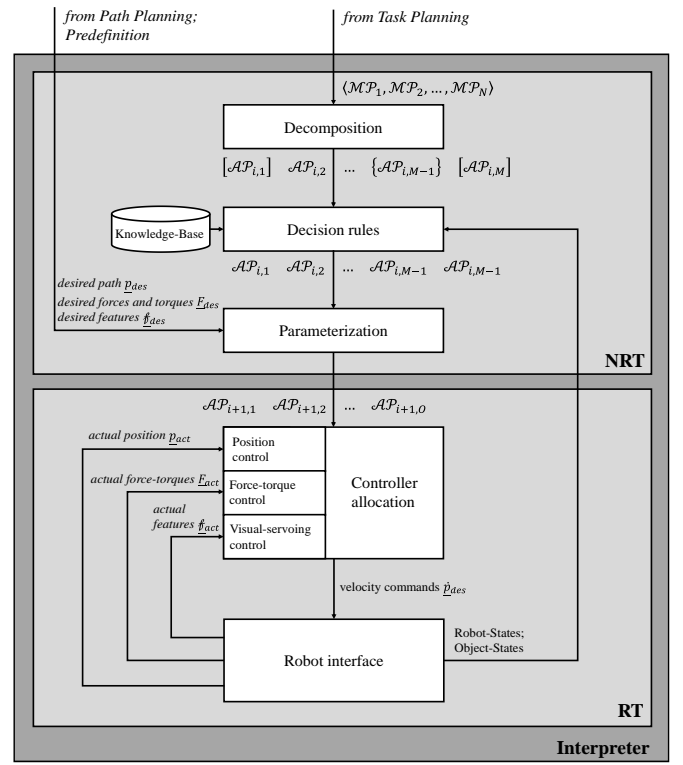

Fig. 4. Interpreter-based concept for the parsing of the plan into a executable skill primitive sequence.

For a better understanding of the decomposition process we will introduce a simple example with a screwing operation, which explains the different steps, see Fig. 5. The task is to disassembly the screw, means there is exactly one manipulation primitive with $twist(screwdriver, S)$. Using the rule (4) the different skill primitives can be derived. Because (9) is $true$ the first step is to get the tool. (14) is $false$, means nothing is executed. The next step is (11) and (12) for the positioning, both $true$. Afterwards the screwing process is executed. This is also done using predefined skill primitives, but to explain the principle, it makes no sense to decompose the process further. Because the put position is predefined, only (11) is used before the screw is put down (13). The release movement (11) and the placement of the screwing unit (10) leads to the termination of the execution.

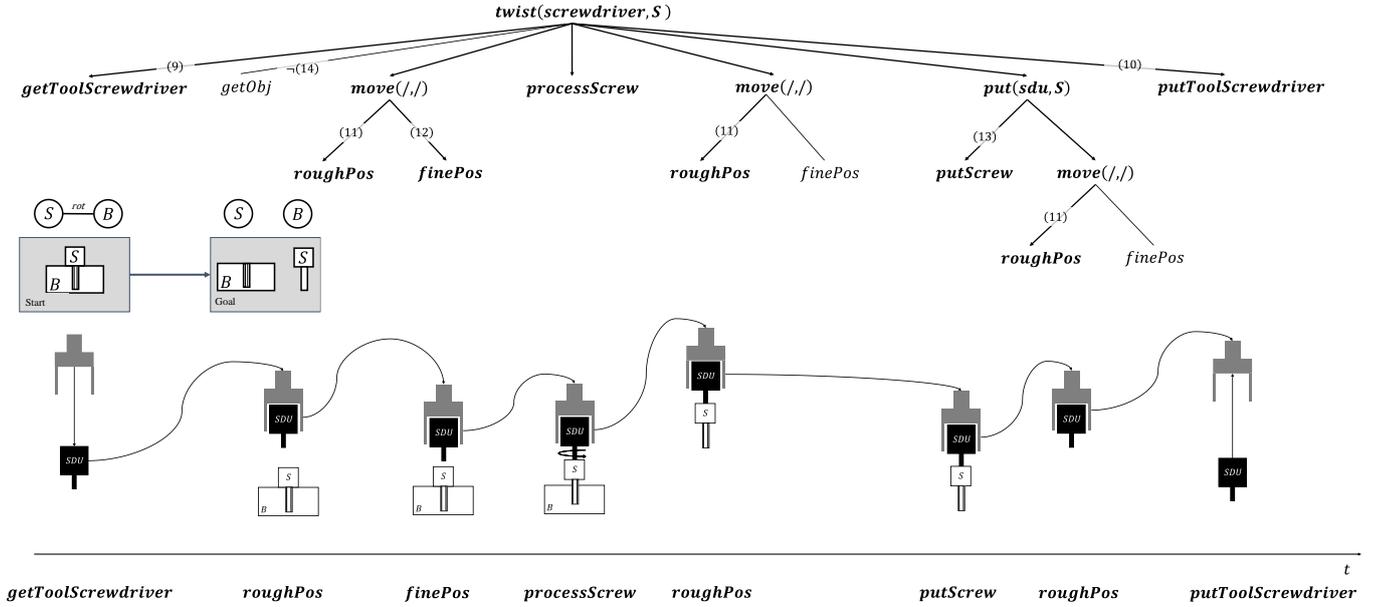

Fig. 5. Example of the decomposition process using a simple screwing operation.

## VI. CARTESIAN VELOCITY CONTROL INTERFACE

The control structure is designed as a switching controller as suggested by the principle of skill primitives [27]. For the possibility to couple the complete framework also to commercial industrial robot controllers, the control output of the switching controller is computed as a desired velocity command, so $\underline{u} \in \dot{\underline{X}}$. The underlying robotic system uses a standard PI-PI cascaded controller designed in the robot joint space. As switching controllers an admittance controller [28] $\underline{R}_A$ is used for force-torque control and an image based visual servoing (IBVS) controller [29] $\underline{R}_{IBVS}$. In addition, a position controller $\underline{R}_p$ is integrated into the switching controller.

Further, the admittance and the IBVS controller will be evaluated to explain the dynamics and the accuracy of the complete system in section VII.

### A. Admittance controller

If $\underline{u}_A \in \dot{\underline{X}}$ (velocity) the admittance control law is given by

$$\underline{u}_A = \underline{C}^{-1} \left( \underline{F}_{des} - \underline{F}_{act} - \underline{M}\, \ddot{\underline{u}}_A - \underline{D}\, \dot{\underline{u}}_A \right). \quad (15)$$

$\underline{C} = diag(c_1 \quad \cdots \quad c_6)$ describes the stiffness, $\underline{D} = diag(d_1 \quad \cdots \quad d_6)$ the damping and $\underline{M} = diag(m_1 \quad \cdots \quad m_6)$ the mass parameters from the controller. The parameters are defined experimentally by $m_i = 5$; $d_i = 250$ and $c_i = 500$ with $i = 1,2,...,6$. $\underline{F}_{des}$ is the desired force and $\underline{F}_{act}$ the measured from our force-torque sensor.

In Fig. 6 left, the robot configuration to evaluate the admittance controller is shown.

Fig. 7 shows the results for a single force-controlled coordinate in the z-direction of the tool frame. For the different contact forces, sufficient reference-variable response is available. The limited dynamics is mainly caused by the available 50 Hz of the main controller frequency.

### B. Image-based visual servoing controller

To do the fine positioning an IBVS controller is applied. For IBVS the standard control law (16), which represents a P-Controller, is used, with $\underline{u}_{IBVS} \in \dot{\underline{X}}$.

Currently, external point features for every component are used for $\underline{\textit{f}}$ to determine the feature Jacobian $\underline{J}_{\textit{f}}$. The P-gain is chosen as $k_{ibvs} = 0.125$.

$$\underline{u}_{IBVS} = k_{ibvs} \cdot \left( \underline{J}_{\textit{f}}^T \, \underline{J}_{\textit{f}} \right)^{-1} \underline{J}_{\textit{f}}^T \cdot \left( \dot{\underline{\textit{f}}}_{des} - \dot{\underline{\textit{f}}}_{act} \right). \quad (16)$$

To evaluate the position accuracy, different start poses are specified according to Fig. 6 right and Fig. 8 top left. In the present case, four features are always considered, so that there is additional redundancy. The controller frequency is reduced by feature extraction. In this example we deal with 20 Hz.

The deviation to the target position for the Cartesian coordinates is 0.05 m. The arithmetic mean of the Cartesian positioning accuracy is 0.0012 m and thus sufficient for the applications in Section VII.

Fig. 8, top right, shows the variation in Cartesian space. Above, the behavior in the image plane is shown respectively for pose four.

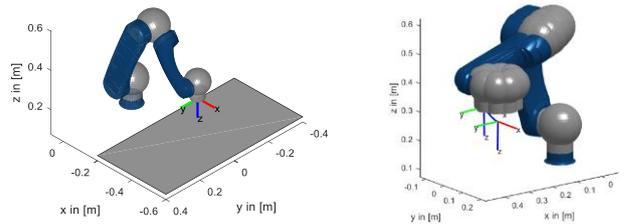

Fig. 6. Robot configuration for the experiments with the admittance controller (left); The different configurations for the experiments with the IBVS-controller. The marked path is for configuration four (right).

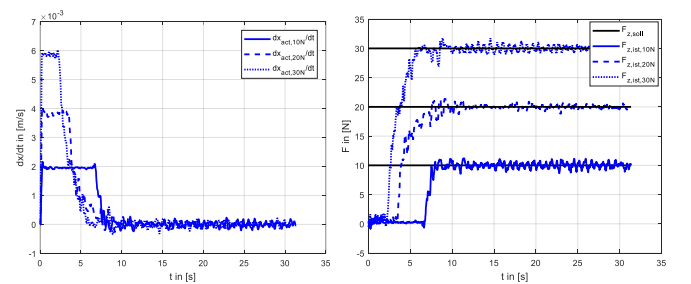

Fig. 7. Results admittance control for $F_{des} = \{10,20,30\}$ in [N]: Controller output (left); Actual and desired forces (right).

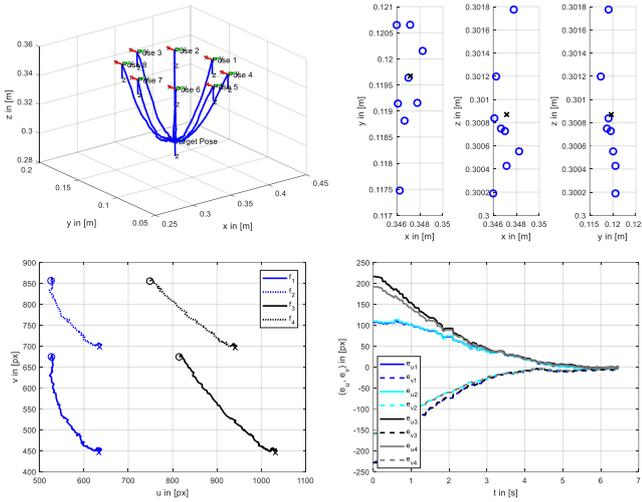

Fig. 8. Results IBVS-controller: Cartesian paths for different start poses (top left); Cartesian variation for the different start poses (top right); trajectory in the u-v image plane for pose four (bottom left); corresponding error (bottom right).

## VII. EXPERIMENTAL RESULTS

Finally, the practical capability of the proposed system is evaluated using four different real-world applications, presented in Fig. 9. The used hardware setup is described in section IV-C. The first application is a peg-in-hole assembly. The panel assembly is a screwed plate for which two screwing tasks have to be performed. For the lubricant exchange the robot must disassemble the screwed cover. The fourth application is a repair task in which a valve must be exchanged. In all experiments, the robot must autonomously perform the disassembly and the assembly task.

### A. Metrics for the experiments

The analysis of the applications described above is done by the following metrics:

- Mean $t_{-,exe}$ and standard deviation $t_{\sigma,exe}$ of the **execution time** for the complete task in $[s]$.
- Time in $[s]$ as mean value $t_{-,path}$ and standard deviation $t_{\sigma,path}$, in which the **robot operates in position control**.
- Time in $[s]$ as mean value $t_{-,vsc}$ and standard deviation $t_{\sigma,vsc}$, in which the **robot operates in visual-servoing control**.
- Time in $[s]$ as mean value $t_{-,ftc}$ and standard deviation $t_{\sigma,ftc}$, in which the **robot operates in force-torque control**.
- Time in $[s]$ as mean value $t_{-,n}$ and standard deviation $t_{\sigma,n}$, in which the **robot operates in non-productive time**. Gripping processes and sensor calibration, in which the robot does not perform any movement, are summarized as non-productive time.
- **Number of manipulation primitives** to be executed $|\mathcal{MP}| \in \mathbb{N}_+$.
- **Success rate** $S \in [0 \dots 1]$ for a correct execution. The success is only positive, if no human-operator is required. This means, that the complete execution is autonomously done by the robot. We distinguish between different errors to evaluate the success rate. The possible error types which affects the success rate are:

  o **Planning error:** The planned actions or the planned trajectory lead to the error.
  o **Sense and control error:** The sensor data and the control used leads to the error. For example, the data from the force-torque sensor is to noisy and the put down position is not detected.
  o **Device error:** The tool used leads to the error. For example, the gripped component is not held due to friction, or the screw is not held by the tool during the screwing process.

All experiments are repeated five times for the evaluation of the success rate. The experiments fail, if the system cannot fully autonomously solve the task. Table II shows the results for the different metrics. Fig. 10 shows also the time measurements for a better presentation of the results.

### B. Discussion of the results

It should be noted beforehand that the overall dynamics of the system is very limited by the low cycle time of the controller. By adjusting the control and the kinematic path parameters, an improvement could be achieved, but this is not in the scope of the present research. The temporal interrelation between the different control methods is investigated instead.

For the peg-in-hole application the force-torque controller is mostly active. Because of different start positions for the joining process and friction effects, the biggest standard deviation in the execution time can be observed.

All other scenarios are mainly position-controlled. The generally long times for force-torque controlled manipulations are, on the one hand, due to the low dynamics of the admittance controller.

With visual-servoing, the dynamics, apart from the known general conditions, are also limited by the time duration to determine the features (e.g. 20 Hz). The standard deviation for image-controlled manipulations can be attributed to light conditions and subsequent shading. Also, the start position varies due to localization errors from object recognition results.

Furthermore, causes should be reviewed for a non-successful task execution. The lubricant exchange task is successfully executed for all five experiments. For the other three applications there is always one experiment that failed. For the peg-in-hole application the joining transition is one time detected to early, because of sensor noise (sense and control error). For the panel and the valve assembly the problem was the removal of the screw from the storage position. This mainly results from the mechanical design of the screwing unit (device error). Another point is the loss of the contact force due to the low dynamics of the admittance controller. This means that the success rate is mostly influenced by a device error (two times). A planning error did not occur, which underlines the suitability of the method for generating skill primitives from symbolic plans.

The different steps of the execution for the valve task are shown in Fig. 11 and Fig. 12 which shows the planned manipulation primitive sequence as well as the created skill primitive sequence (only the disassembly steps) with the method from section V. To solve the repair task, the robot must disassemble two screws and an air hose, compare the manipulation primitive sequence in Fig. 11.

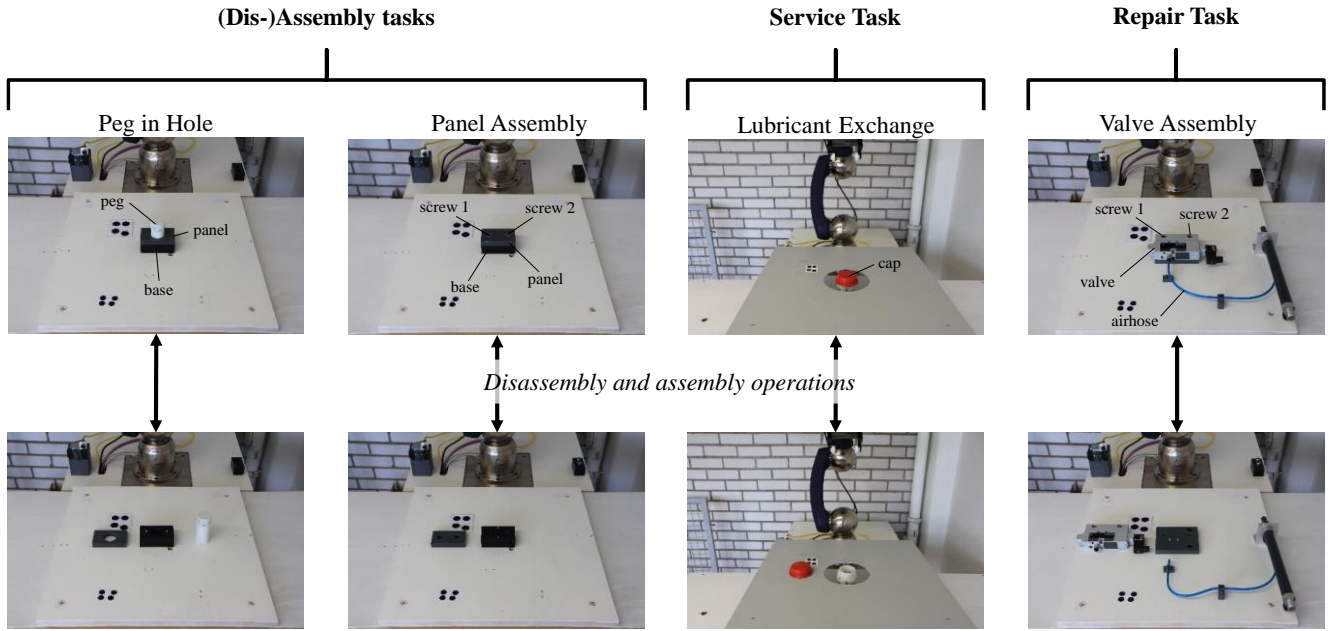

Fig. 9. Used application scenarios for the evaluation of the manipulation capabilities.

TABLE II. EXPERIMENTAL RESULTS FOR ALL APPLICATION SCENARIOS.

|  | *application 1- peg in hole* | *application 2- panel assembly* | *application 3- cooling lubricant* | *application 4- valve assembly* |
|---|---|---|---|---|
| $t_{-,exe}$ in $[s]$ | 271.456 | 466.574 | 505.517 | 544.385 |
| $t_{-,path}$ in $[s]$ | 96.338 | 321.079 | 448.965 | 363.753 |
| $t_{-,vsc}$ in $[s]$ | 14.876 | 21.926 | 5.369 | 35.473 |
| $t_{-,ftc}$ in $[s]$ | 137.145 | 93.632 | 31.125 | 107.796 |
| $t_{-,n}$ in $[s]$ | 23.097 | 29.937 | 20.057 | 37.362 |
| $t_{\sigma,exe}$ in $[s]$ | 14.116 | 8.770 | 3.326 | 6.264 |
| $t_{\sigma,path}$ in $[s]$ | 0.674 | 1.982 | 0.611 | 3.887 |
| $t_{\sigma,vsc}$ in $[s]$ | 1.453 | 4.419 | 0.354 | 7.882 |
| $t_{\sigma,ftc}$ in $[s]$ | 14.041 | 7.320 | 2.716 | 9.583 |
| $t_{\sigma,neben}$ in $[s]$ | 0.166 | 0.049 | 0.256 | 0.283 |
| $|\mathcal{MP}| \in \mathbb{N}_+$ | 4 | 10 | 2 | 12 |
| $S \in [0 \dots 1]$ | 0.8 | 0.8 | 1 | 0.8 |

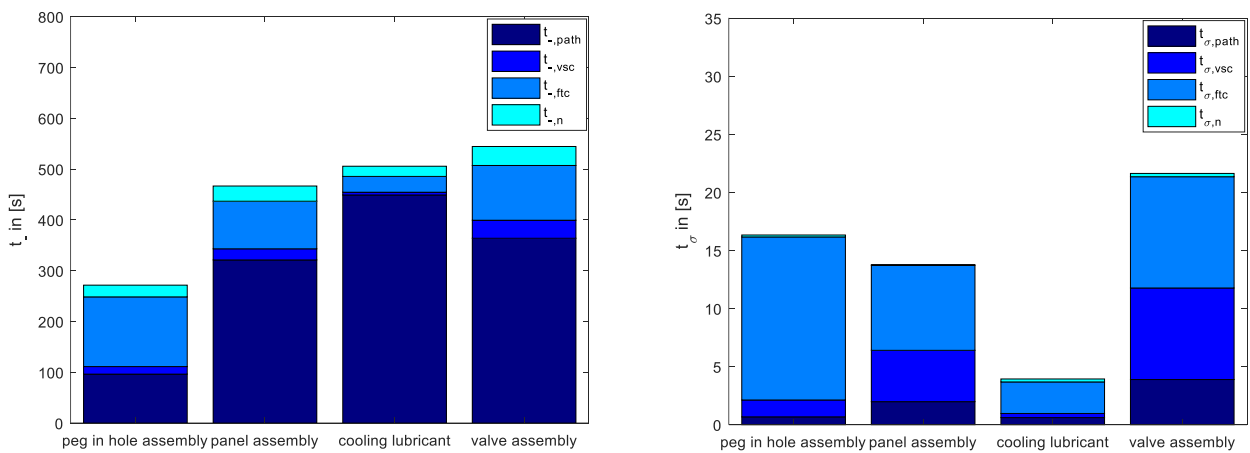

Fig. 10. Execution time for the different applications. Average time (left); standard deviation time (right).

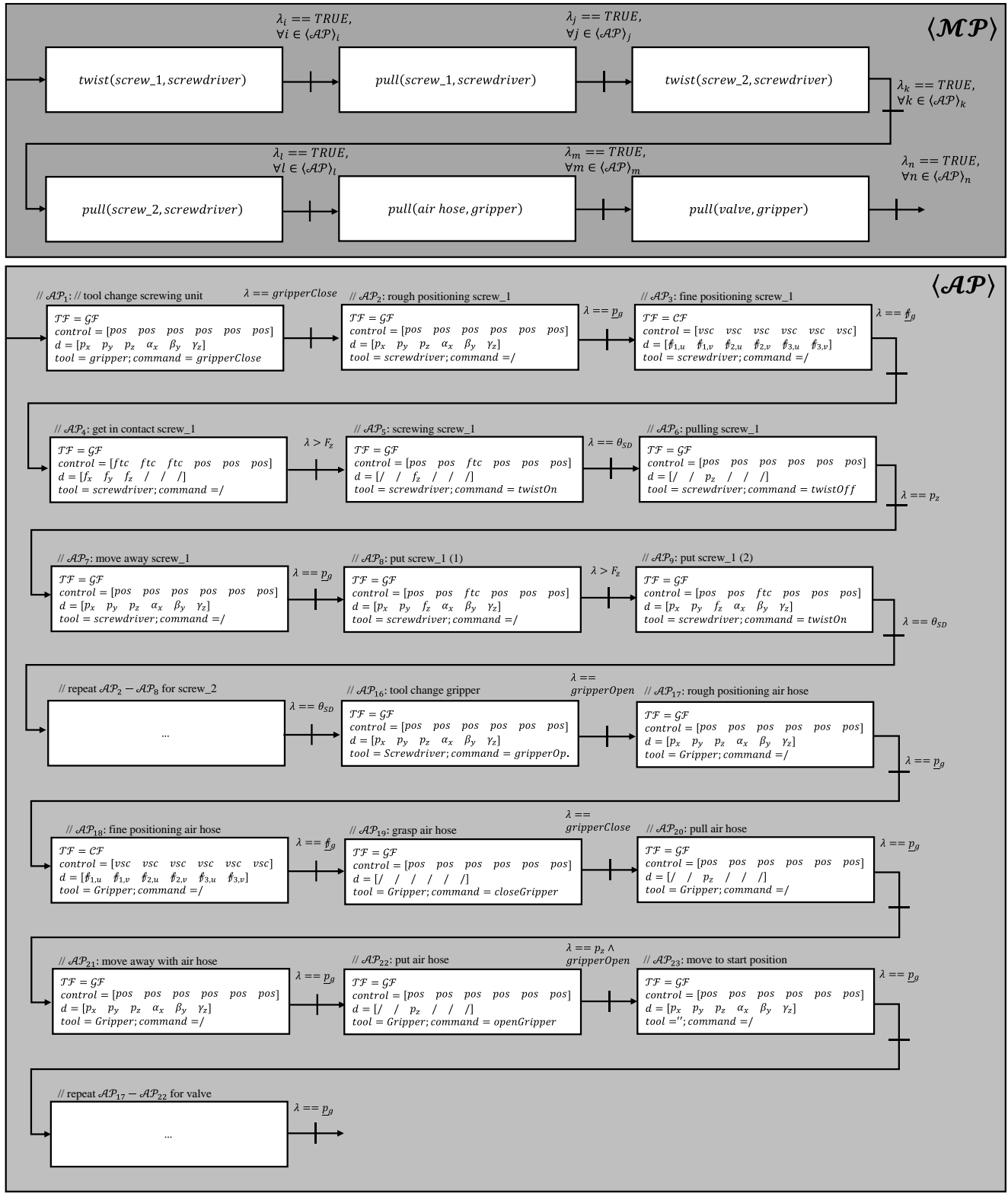

Fig. 11. Created skill primitve sequence from the symbolic plan for the valve assembly. The $\langle \mathcal{MP} \rangle$-sequence shows the symbolic plan from the task planner. The $\langle \mathcal{AP} \rangle$-sequence above is the output from the plan decomposition using the decision rules (control program generation). In Fig. 11 is the corresponding execution through the used robotic system.

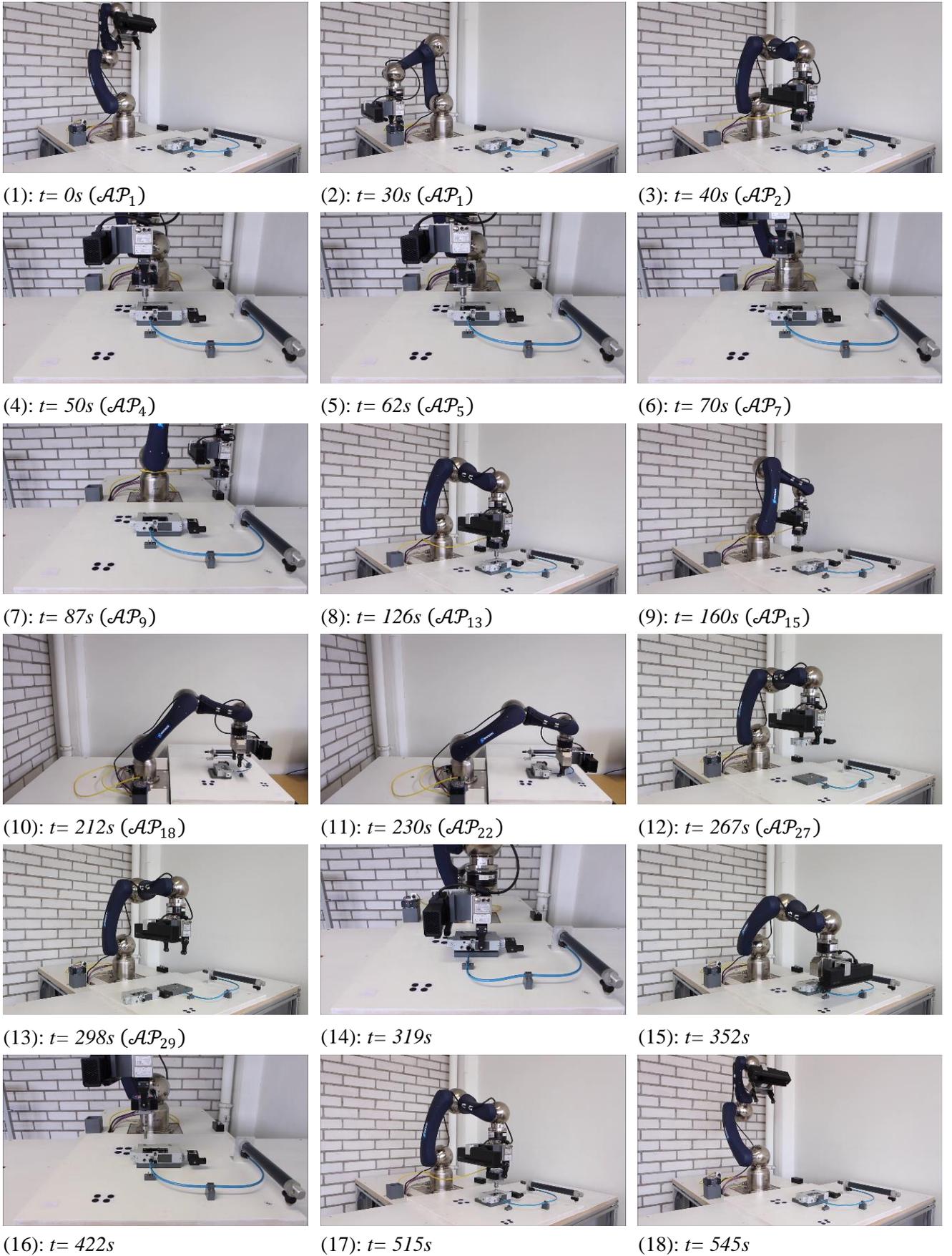

Fig. 12. Execution of the valve disassembly and assembly for the created plan described in manipulation primitives and the corresponding skill primitive sequence in Fig. 10. In the steps (1) - (13) the robot execute the disassembly steps. Step (13) shows the final disassembly state. From (14) – (18) the assembly steps are shown, which is the inverse plan showed in Fig. 11.

The first skill primitive controls a tool change from the gripper to the screwing unit (Fig. 12 (1)-(2)). For dismantling the screw at first, a rough positioning is done using the pose coming from object detection (Fig. 12 (3)). Fine positioning is then carried out using the IBVS controller (Fig. 12 (4)). After that, the unscrewing process takes place and the storage of the screw in a predefined position (Fig. 12 (5)-(7)). The same procedure is done for the second screw (Fig. 12 (8)-(9)). After executing a tool change, the air hose disassembly starts with the same approach, using rough and fine positioning and the process-defined disassembly function (Fig. 12 (10)-(11)). In the last disassembly step, the robot puts the valve on the table (Fig. 12 (12)-(13)). In the remaining steps the valve is assembled (Fig. 12 (14)-(18)) using the inverse sequence.

## VIII. CONCLUSION AND FUTURE WORK

This paper proposes methods for the planning and execution of complex manipulation tasks for maintenance automation. We present the required components and skills, from planning to execution, to solve this problem for a number of real-world applications. Through the integration of sensor data into the planning process it is possible to incorporate environmental uncertainties. The concept of high-level plans combined with a sampling-based pre-processing approach enables a fast task planning in $O(n \cdot log(n))$. By planning general manipulation sequences, a flexible parsing into skill primitives is allowed during runtime. This concept allows the adaption of the plan during operation time. We had demonstrated the applicability of the presented methods on four different examples, which demonstrates the effectiveness of the proposed approach.

Future work will focus on dual-arm manipulation. This also requires the adaption of the task planning process, because now only linear, monotonous plans (no sub-assemblies compare [30]) can be created. Another big problem is uncertainty during task execution. What are alternative strategies if a component cannot be disassembled due to wear or corrosion? How can an alternative plan be generated? All these points are of great relevance for the application of modern robotic systems into everyday industrial life, since, in addition to high functionality, a high failure safety and reliability of the system must also be ensured.

Solving these problems can help us to establish complex robotic systems to support the human colleague and to make manufacturing systems more efficient.